# Neural Networks in 3D medical scan visualization

Dženan Zukić, Dipl. Ing., dzenanz@gmail.com

Hotonj II br. 52, 71322 Vogošća, Bosnia and Herzegovina

University of Sarajevo / Faculty of Electrical Engineering / Computer Science Department

Andreas Elsner, Dipl. Inf., trinet@uni-paderborn.de

Warburger Strasse 100, D-33098 Paderborn, Germany; Office: E3.118

University of Paderborn / Computer Sci. / Computer Graphics / Visualization and Image Processing

Zikrija Avdagić, Prof. Dr., zikrija.avdagic@etf.unsa.ba

ETF, Zmaja od Bosne bb - Kampus, 71000 Sarajevo, Bosnia and Herzegovina

University of Sarajevo / Electrical Engineering / Computer Science / Artificial Intelligence

Gitta Domik, Prof. Dr., domik@uni-paderborn.de

Warburger Strasse 100, D-33098 Paderborn, Germany

University of Paderborn / Department of Computer Science / Research Group Computer Graphics

## Abstract

For medical volume visualization, one of the most important tasks is to reveal clinically relevant details from the 3D scan (CT, MRI …), e.g. the coronary arteries, without obscuring them with less significant parts. These volume datasets contain different materials which are difficult to extract and visualize with 1D transfer functions based solely on the attenuation coefficient. Multi-dimensional transfer functions allow a much more precise classification of data which makes it easier to separate different surfaces from each other. Unfortunately, setting up multi-dimensional transfer functions can become a fairly complex task, generally accomplished by trial and error.

This paper explains neural networks, and then presents an efficient way to speed up visualization process by semi-automatic transfer function generation. We describe how to use neural networks to detect distinctive features shown in the 2D histogram of the volume data and how to use this information for data classification.

# Introduction

For visualization and analysis of CT data (or any other 3D medical scan, like MRI or PET), the key advantage of direct volume rendering is the potential to show the three dimensional structure of a feature of interest, rather than just a small part of the data by cutting plane. This helps the viewer's perception to find the relative 3D positions of the object components and makes it easier to detect and understand complex phenomena like coronary stenosis for diagnostic and operation planning [9].

One of the most basic requirements of good volume visualization is a sophisticated way to extract data that is actually of interest, while suppressing insignificant parts. A lot of progress has been made in the development of transfer functions as a feature classifier for volume data. The role of transfer functions is basically to assign specific visual attributes like color and opacity to the features of interest. The simplest approach would be to use the voxel data as the only variable to which these visual attributes are assigned. But this method fails in most cases to omit interference of other anatomical structures, because datasets created from CT scans contain a combination of different materials and tissue types with overlapping boundaries. These structures may contain the same range of data values making it impossible for 1D transfer functions to differentiate between them.

Instead of classifying the volume data only by the scalar value of its voxels, multi-dimensional transfer functions solve this problem by adding additional dimensions to the transfer function domain. By using a classification based on a combination of properties, different features can then be visualized separately from each other. Therefore two dimensional transfer functions can be used for the classification process which, in addition to the attenuation coefficient, also takes the corresponding gradient magnitude into account. At this time there exist different approaches to the design of transfer functions, which become more widely propagated [1], [2]. One of the most effective ways is the use of manually editable, multi-dimensional transfer functions [3]. For this purpose, the software "VolumeStudio" was developed capable of visualizing both modalities.

Unfortunately, specifying two-dimensional transfer functions can become a fairly complex and time-consuming task. First, the user must handle additional parameters for the second dimension, thus making the editing process more difficult. Second, the user must set the transfer function in the rather abstract transfer function domain while observing the result in the rendered image (spatial domain). The situation is made worse by the fact that small changes to these parameters tend sometimes to cause large and unpredictable changes to the visualization. Thus, to avoid editing based completely on trial and error, it is necessary to identify features of interest in the transfer function domain first, requiring a good understanding of the data values and their relevance for a certain feature. Additionally, the manual assignment of optical properties might be non-standardized but rather influenced by the personal taste of the user. This can cause misinterpretation of the data if the visualization is later on reviewed by others. Also, this often leads to results which are hardly reproducible and hence not suitable for clinical practice. Finally, radiologist have a limited time allowed for diagnostics per patient, which makes two-dimensional transfer functions impractical to use without proper guidance through the process or a semi-automatic transfer function generation.

This paper presents a new approach for two-dimensional transfer function generation based on neural networks. Although this technique is flexible enough for classification of different types of CT dataset, in this paper we focus on heart scan visualization to detect coronary diseases. As histograms of same scan type (e.g. heart scans) have similar structures (same basic shape), neural networks can be trained to position filters on features of interest according to the diagnostic target.

Some of the related work can be found in [4] and [5].

## Volume rendering

For the volume rendering of scalar volume data like CT scans, different approaches exist. Texture based techniques have proved superior, combining high quality images and interactive frame rates. These approaches take advantage of the hardware support of bilinear and trilinear interpolation provided by modern graphic cards, making high quality visualization available on low cost commercial personal computers. For these approaches the dataset is stored in the graphics hardware texture memory first. If the size of the dataset exceeds the available memory, bricking can be used to render the data in multiple steps. The dataset is then sampled, using hardware interpolation.

2D texture-based approaches use three copies of the volume data which resides in texture memory. Each copy contains a fixed number of slices along a major axis of the dataset which will be addressed depending on the current view direction. After bilinear interpolation, the values of the slices will then be classified through a lookup table, rendered as a planar polygon and blended into the image plane. This method often suffers from artifacts caused by the fixed number of slices and their static alignment along the major axes. Alternatively, hardware extensions can be used for intermediate slices along the slice axis to achieve better visual quality.

Modern graphics cards support 3D texture mapping which allows storing the whole dataset in one 3D texture. It is then possible to sample view-aligned slices using trilinear interpolation. This approach avoids the artifacts which occur when 2D texture-based techniques switch between the orthogonal slice stacks and allows an arbitrary sample rate, which results in an overall better image quality. Also, no additional copies of the dataset are necessary, lowering the requirements of texture memory.

## Transfer functions

The value of volume rendering in medicine, science and other fields of application not only depends on the visual quality achieved with the rendering process itself, the most important task is to find a good classification technique that captures the features of interest while suppressing insignificant parts. As mentioned above, classification can be achieved by transfer functions, which assign renderable optical properties like color and opacity to the values of the dataset.

2D transfer functions classify the volume not just on the data values but on a combination of different properties and therefore the boundaries of different structures in the dataset can be better isolated as with 1D transfer functions. This is because the structures and tissue types which are to be separated might lie within the same interval, making 1D transfer functions unable to render them in isolation.

Kniss et al. presented a method for manual multi-dimensional transfer function generation based on the data values and its derivatives [3]. The gradient is useful as an additional criterion for classification since it discriminates between homogenous regions inside a structure and regions of change at the boundaries. Also, the gradient can be used to apply illumination to the volume visualization which improves depth perception. The manual editing is performed in a visual editor, which illustrates the distribution of tuples of attenuation coefficient and gradient magnitude of the dataset in a joint histogram. The attenuation coefficient is shown on the x-axis, the gradient magnitude on the y-axis. An example is given in Figure 1.

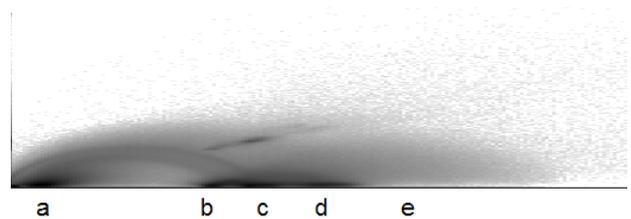

*Figure 1 2D joint histogram of attenuation coefficient versus gradient magnitude. Regions a) – e) identify different materials.*

Homogeneous regions appear as circular spots at the bottom of the histogram, as their gradient magnitude is very low. Five basic regions can be identified: a) air, b) soft tissue like fat and skin, c) muscles, d) blood (with contrast agent) and e) bones. Arches between these spots represent the boundaries of these regions.

To create a new transfer function, the user places filters inside the histogram, shown as rectangular areas. Each filter assigns color and opacity values to the voxels of the dataset which are represented by the tuples of attenuation coefficient and gradient in the histogram inside the defined area. The filter size and position can be changed, also its color and opacity distribution. Also, the shape of the filter kernel can be altered between Gauss and sine shape. If multiple filters are set, their color and opacity values are blended together. The above editing can be done interactively, and the user can decide by observing the visualization in the spatial domain if the current setup is appropriate or not. Through this feedback it is also possible to identify which regions of the transfer function correspondents to the features of the dataset.

Figure 2 shows a volume rendering of a CT scan of the heart and the transfer functions used. It consists of two gauss filters: The first one colored in yellow is located between the regions c) and d) (compare Figure 1) to visualize the myocardial muscle and the coronaries (by contrast agent). The second one resides at the top of the first filter, enhancing the contrast between myocard and coronaries by coloring the properties that represent the boundaries of the contrast agent in red.

For an experienced user, the distinctive features of the distribution shown in the histogram provide useful information about the features metrics, thereby guiding the transfer function generation. But even with these hints, this is a time-consuming iterative process. The user has to explore the dataset by defining filters and move them to possible interesting locations on the histogram. Once a feature of interest is identified, the parameters for the filter size, location, filter kernel shape, opacity and color have to be optimized to match with the user's needs until all features of interest are made visible.

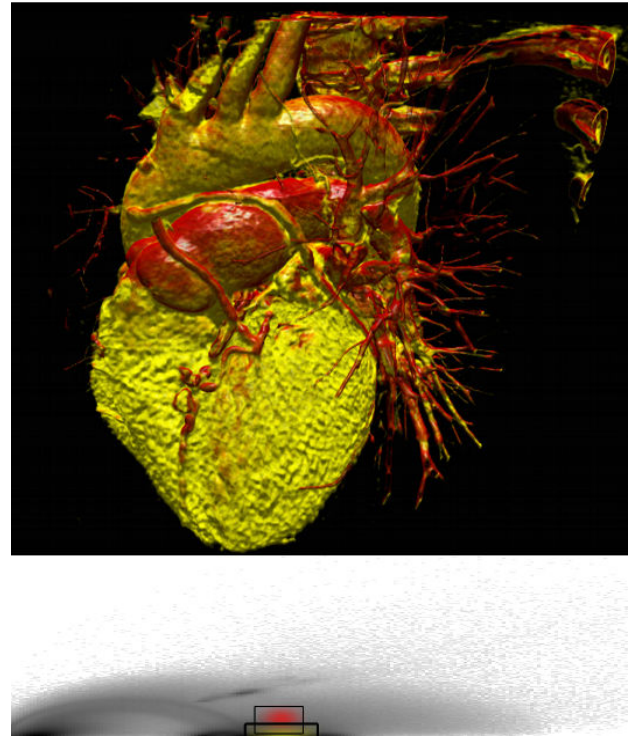

*Figure 2 Volume rendering of a thoracic CT scan classified with a 2D transfer function*

## Neural networks

A neural network is a structure involving weighted interconnections among *neurons* (which are most often nonlinear scalar transformations). A neuron is structured to process multiple inputs, usually including the unity bias, in a nonlinear way, producing a single output. Specifically, all inputs to a neuron are first augmented by multiplicative weights. These weighted inputs are summed and then transformed via a nonlinear *activation function*. The weights are sometimes referred to as *synaptic strengths*. The general purpose of the *Neural Networks* can be described to be function approximation.

When input data originates from a function with real-valued outputs over a continuous range, the neural network is said to perform a function approximation. An example of an approximation problem is when we control some process parameter by calculating a value of certain (complex) function. Instead, we could make a neural network that approximates that function, and a neural network calculates output very quickly.

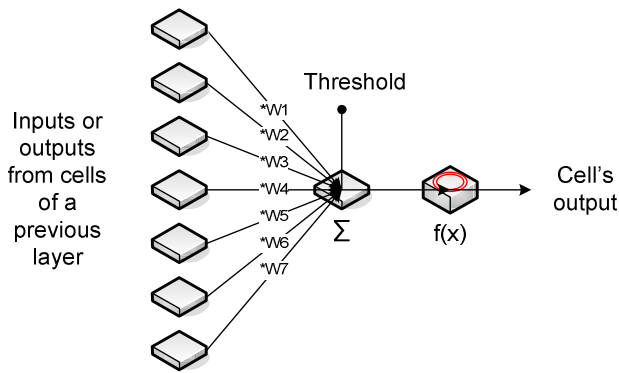

*Figure 3 Neural cell - neuron*

The feed-forward, back-propagation network architecture was developed in the early seventies. The initial idea is credited to various individuals, most prominently D.B Parker, P.J Werbos, G.E Hinton, R.J Williams, and D.E Rumelhart.

One of the most commonly used networks is the multilayer feed forward network, also called multi-layer preceptron. It has widespread interesting applications. We have isolated this specific network type as the neural network we have used ourselves.

Feed-forward networks are advantageous as they are the fastest models to execute, and are universal function approximators. One major disadvantage of this network type is that no fast and reliable training algorithm has yet been designed and therefore can be extremely slow to train. Thus, multilayer feed-forward networks should be chosen if rapid execution rates are required, but slow learning rates are not a problem.

Feed-forward networks usually consist of three or four layers in which the neurons are logically arranged. The first and last layers are the input and output layers respectively and there are usually one or more hidden layers in between them. Research indicates that a minimum of three layers is required to solve complex problems [6]. The term feed-forward means that the information is only allowed to "travel" in one direction (there are no loops in networks). Furthermore, this means that the output of one layer becomes the input of the next layer, and so on. In order for this to happen, each layer is fully connected to next layer (each neuron is connected by a weight to a neuron in the next layer.

Depending on implementation, input layer can just distribute data to second layer, or it can subject inputs to threshold (if implemented in network) and activation function.

The output of each neuron (except of those in input layer) is computed like: $y_j = f(\theta + \sum_i w_{ij} * y_i)$, i –previous layer, j – current layer, y – output, f – activation function, $\theta$ – threshold (if implemented in network). For input layer: $y_i = f(\theta + x_i)$.

In order to train the neural network, sets of known input-output data points must be assembled. In other words, the neural network is trained *by example*; much like a small child learns to speak.

The most common and widely used algorithm for multi-layer feed-forward neural networks is the back-propagation algorithm.

The algorithm starts by comparing actual output of the network for the presented input with the desired output. The difference is called output error, and we try to minimize it [8]. For each output in the output layer error delta is calculated: $\delta_f = y'_f * (desired_f - y_f)$, where $y'_f$ is first derivative of the activation function in that output neuron for the point of its output.

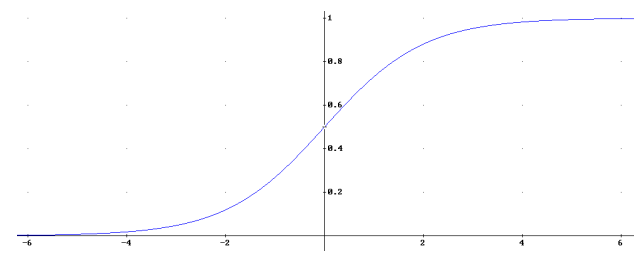

*Figure 4 Logistic activation function*

The need to calculate derivatives in the process of backpropagation explains the usual choice of the activation function: logistic ($\frac{1}{1+e^{-x}}$) or hyperbolic tangent ($\frac{e^{2x}-1}{e^{2x}+1}$). The main reason is that derivatives of these functions can be calculated using the value of the function. For logistic function, which was used in our neural network, equality holds: $y' = y * (1 - y)$.

Activation functions have to be differentiable in order to apply backpropagation.

After we have error deltas in the output layer, we can update weights leading into it, using the following rule: $w_{ij} = w_{ij} + \eta * \delta_j * y_i$, where $\eta$ is learning rate (small value, usually between 0.05 and 0.3), $\delta_j$ is error delta of the neuron weight is leading into, $y_i$ is output of the neuron weight is coming from (output produced in the feed-forward process). This rule is subsequently used for all weights. Weights can be updated individually (called on-line learning) or in groups (called batch learning). In batch learning, we calculate weight deltas for all samples, and then combine them in one single update. Batch update is usually better, as it lowers oscillations of the weights.

For all layers except output, error deltas are calculated using rule: $\delta_i = y'_i * \sum_j w_{ij} * \delta_j$, i – current layer, j – next layer. We proceed backwards to the first layer calculating error deltas and updating weights. When the weights between 1$^{st}$ and 2$^{nd}$ layer are updated, backpropagation is finished.

The training process is repeated many times (epochs) until satisfactory results are obtained. Training can stop when the error obtained is less than a certain limit, or we have reached some pre-set maximum number of training epochs [7].

It is important to say that "over-training" of a network should be avoided, as it lowers predictive abilities of the network, as it is said that network learns "details of the training set". Examples that the network is unfamiliar with, form what is known as the validation set, which tests the network's capabilities before it is implemented for use.

## Solution of the problem

Knowing the properties of 2D histograms and 2D transfer functions on one hand, and having some experience in neural networks on the other, we conceived a solution to the problem.

As stated in transfer functions section, the 2D histogram showing the distribution of tuples of attenuation coefficient and gradient magnitude of a heart dataset contains distinctive features which can be used to guide the transfer function setup. These features consist of circular spots at the bottom of the histogram representing homogeneous materials and arches which define material boundaries. Hence, the poison and size of a filter setup for a 2D transfer function depends on those patterns.

Given as an input, the histogram can be used to train a neural network for pattern recognition. Therefore the user creates filter setups for a training set manually according to the diagnostic target. The network is then trained to associate outputs (filters) with input patterns in the histogram. This time consuming step has only to be performed once and can be done outside clinical practice. Once the network is properly trained, it can be used to create an appropriate filter setup automatically.

The 2D histogram is basically a grayscale image with dimensions 256*256. An input of this size would require a significant amount of memory for storage (16MB just for weights in case of 64 neurons in 2$^{nd}$ layer). Also, training of such a network would be slow, and its generalization abilities would be presumably low.

Therefore, as a preprocessing step, the input to the neural network must be reduced to data that is relevant for the pattern recognition. First, we removed those parts of the histogram that contain no data at all. Since the size of the used part in the histogram varies from dataset to dataset, we estimated a maximum size based on the training set and use this as a fixed value for all datasets of the same type. For the heart CT scans used to evaluate our approach it is sufficient to remove the upper half of the histogram and only take the lower one into account. Second, we downscaled the remaining histogram by a factor of 4. This reduces the number of inputs to the neural network to just 2048. Furthermore, the downscaling of the image smoothes out parts of the histogram which lie outside the distinctive features required for pattern recognition. Since these parts consists of tuples of attenuation coefficient and gradient magnitude which have only a few voxels of the dataset assigned to them, they appear to the neural network as noise. Also, these parts vary a lot between different datasets. As this affects the learning rate, with noise removed and image size reduced, the neural network will learn more easily and will have better generalization abilities.

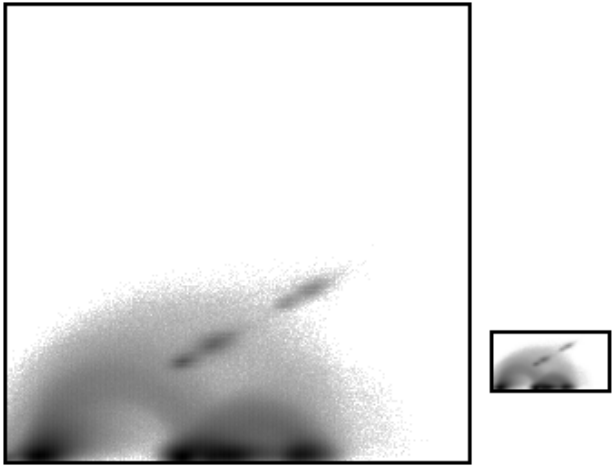

*Figure 5 Original and reduced histogram*

As two gauss filters are usually used to visualize heart and its arteries, we decided that output of our network would be positions and sizes of those gauss filters. Hence, number of outputs is 8 (xpos1, ypos1, xsize1, ysize1, xpos2, ypos2, xsize2, ysize2).

That leaves some variability for layers in between. We started with 1 hidden layer with 64 neurons in it. We worked with this architecture throughout software development until final training and testing, which is when we did some experimentation. We reduced size of the hidden layer first to 32 and then to 16, and noticed no degradation in results. We kept 16 neurons in hidden layer. We did not experiment with more than 1 hidden layer (as there was no need for it).

Trying to see how many samples need to be used for training in order for neural network to be useful, we did some experimentation. For 12 samples, we manually determined positions. Then 2 samples we marked as control samples (validation samples), and other 10 were used for training. We created 5 neural networks, first one trained with 2 samples, second one with 4 samples and fifth one with 10 samples, and on all of these networks we used 2 control samples to check for error. On Figure 6, there are 2 data series, one showing error of networks on training data, and the other errors on test data. For all networks except first one (the one trained with only 2 samples), mean square error is lower on test set, than on training set. This is unusual, but can be explained with fact that positions that we manually provided for networks, were not all that similar.

It is quite clear that even small number of training samples produces good results. In our measurements, networks trained on 6, 8, and 10 samples provide nearly the same results as network trained with just 4 samples. This can be explained by the fact that histograms have so typical shape, so just 4 training samples suffice for good recognition, and all knowledge gained by additional training is annihilated by "overfitting", so training the network beyond basic needs achieves very little effect.

Also interesting is that training MSE (mean square error) jumps on network trained with 4 samples, and then gradually decreases with increased number of training samples. This can be explained with assumption that either on $3^{rd}$ or $4^{th}$ sample training data was "radically" different from the others, so network could not easily minimize that errors that its oddity produces. As the number of samples increase, relative influence of that sample is reduced and MSE is lowered.

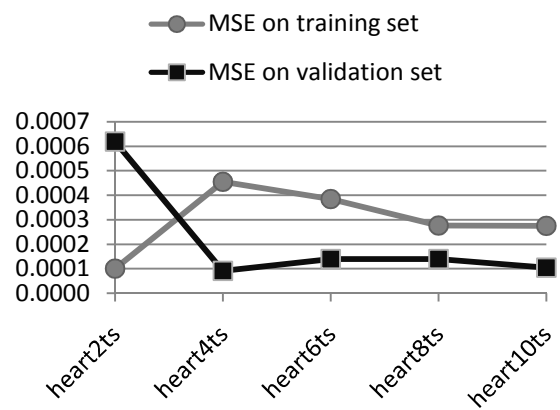

*Figure 6 Mean square errors on networks trained with varying number of train samples*

The neural network software built into VolumeStudio enables to additionally train existing neural network, and stand-alone training tool has also be created (with more features for training than built-in function). This enables neural network to be retrained, or created from scratch on new dataset. Creating new neural network also enables creating specialized networks for other specific body scans, like head or the whole body, for example (we did not have enough samples of those types to experiment with it ourselves).

## Conclusion and outlook

The time spent on positioning the filters has been cut down from 1-3 minutes to approximately 10-30 seconds needed for fine tuning of the parameters after automatic filter generation, giving doctors more time to analyze the data. Also, neural networks kick-start usefulness of VolumeStudio for new users.

A number of things could have been done differently. First, histogram image downscaling could be by factor of 2, not 4. We did not change that, because the results are satisfactory as it is done now. We could have experimented with different number of layers, to see what results it would give.

Also, as it is now implemented in software, the user has to manually select an appropriate neural network to create the filters by the type of dataset (e.g. heart, head or whole body scan) that has been loaded.

One approach to automate this too, is to use an additional network to classify input samples into type categories. This network has to have as many outputs as there are different networks for different data types. When the user loads new scan, this data classification network is used to determine type of scan and after that, based on the output of the classification network the appropriate network for filter positioning is chosen. This approach, however, has the small drawback that whenever you add a network for new scan type, you have to change architecture of the classifier by adding an additional output and subsequently re-train it.

## References

[1] Engel, K., Hadwiger, M., Kniss, J., Rezk-Salama, C.: Real-Time Volume Graphics. Eurographics Tutorial (2006)

[2] Engel, K., Hadwiger, M., Kniss, J., Rezk-Salama, C., Weiskopf, D.: Real-Time Volume Graphics. AK Peters, Ltd, Wellesley, Massachusetts (2006)

[3] J. Kniss, G. Kindlmann, C. Hansen. Multi-Dimensional Transfer Functions for Interactive Volume Rendering. IEEE Transactions on Visualization and Computer Graphics, 8(3): 270-285, July, 2002

[4] J. Marks, B. Abdalman, P. Beardsley, W. Freeman, S. Gibson, J. Hodgins, T. Kang, B. Mirtich, H. Pfister, W. Ruml, K. Ryall, J. Seims, S. Shieber. Design Galleries: A General Approach to Setting Parameters for Computer Graphics and Animation. In Proceedings SIGGRAPH, 1997

[5] G. Kindlmann, J. W. Durkin. Semi-Automatic Generation of Transfer Functions for Direct Volume Rendering. ACM Symp. On Volume Visualization, 1998

[6] Zikrija Avdagić, *Artificial Intelligence & Fuzzy-Neuro-Genetic*, Grafoart Sarajevo, 2003.

[7] Bernard Willers and Sep Vrba, „Artificial Neural Networks: Emulating the Operation of the Human Brain", August 2000, http://library.thinkquest.org/C007395/.

[8] D. Rumelhart, G. Hinton and R. Williams, „Learning representations by back-propagating errors", Nature, Vol. 323, pp. 533 – 536, October 1986.

[9] Brochure „Blick in das Herz", Deutschen Gesellschaft für Nuklearmedizin